\documentclass{nle}

\newif\ifjnle
\jnlefalse

\usepackage{adjustbox}
\usepackage{blindtext}
\usepackage[justification=centering,font=small]{caption}
\usepackage{comment}
\usepackage[T1]{fontenc}
\usepackage{graphicx}
\usepackage{multirow}
\usepackage{nicematrix}
\usepackage{quoting}
\usepackage{tabularx}
\usepackage{url}

\graphicspath{{./img/}}

\newcommand{\todo}[1]{\ignorespaces}

\newcommand{\LP}{LivePerson\xspace}
\newcommand{\lego}{LEGOv2\xspace}

\renewcommand{\bot}{dialog system\xspace} 
\newcommand{\bots}{dialog systems\xspace}

\begin{document}

\title{Actionable Conversational Quality Indicators for Improving Task-Oriented Dialog Systems}

\begin{authgrp}
\author{Michael Higgins}
\author{Dominic Widdows}
\author{Chris Brew}
\author{Gwen Christian}
\author{Andrew Maurer}
\author{Matthew Dunn}
\author{Sujit Mathi}
\author{Akshay Hazare}
\author{George Bonev}
\author{Beth Ann Hockey}
\author{Kristen Howell}
\author{\ Joe Bradley}
  \affiliation{LivePerson Inc. \\ 
  \email{mhiggins | dwiddows | cbrew | amaurer | gchristian | mdunn | smathi | ahazare | gbonev | bhockey | khowell | jbradley @liveperson.com}}
\end{authgrp}

\begin{abstract}

Automatic dialog systems have become a mainstream part of online customer service. Many such systems are built, maintained, and improved
by customer service specialists, rather than dialog systems engineers and computer programmers.
As conversations between people and machines become commonplace, it is critical to understand what is working, what is not, 
and what actions can be taken to reduce the frequency of inappropriate system responses. These analyses and recommendations need to
be presented in terms that directly reflect the user experience rather than the internal dialog processing.

This paper introduces and explains the use of Actionable Conversational Quality Indicators (ACQIs), which are used both to recognize
parts of dialogs that can be improved, and to recommend how to improve them. This combines benefits of previous approaches, some of which
have focused on producing dialog quality scoring while others have sought to categorize the 
types of errors the dialog system is making.

We demonstrate the effectiveness of using ACQIs 
on \LP 
internal dialog systems used in commercial customer service applications,
and on the publicly available CMU \lego conversational dataset \citep{Raux2005LetsGP}.
We report on the annotation and analysis of conversational 
datasets showing which ACQIs are important to fix in various situations.

The annotated datasets are then used to build a predictive model which uses
a turn-based vector embedding of the message texts and achieves a
79\% weighted average f1-measure at the task of finding the correct ACQI for a given conversation. 
We predict that if such a model worked perfectly, 
the range of potential improvement actions
a bot-builder must consider at each turn could be reduced by an average of 81\%.

\end{abstract}

\maketitle

\ifjnle
\else
    \thispagestyle{empty}
\fi

\section{Introduction and Outline}

Customer service \bots have become widely-used in many everyday settings, but still make frustrating errors that are sometimes obvious to a human 
--- for example, failing to
understand a customer's message and asking an irrelevant question as a result. 
In practice, tuning these systems to limit these behaviors is an expensive and time-consuming art.
This paper describes the design, implementation
and early results of an approach to improving overall \bot quality by recognizing and addressing such individual failures. This is done by 
combining individual Actionable Conversational Quality Indicators (ACQIs) with a running Interaction Quality score (IQ),
to show which problems were identified, what steps were taken to fix them, and how these changes affected an overall assessment of the user experience. IQ is a standard conversational quality measure developed by \cite{schmitt2015interaction}. 
ACQIs are designed so that each conversational quality indicator has associated recommended actions, and are introduced in this paper.

The paper starts by explaining some of the background on how task-oriented \bots are built and maintained (Section \ref{sec:conv_ai_systems}): 
while there are several online tools that support this, many readers may be unfamiliar with their use.
Section \ref{sec:data} introduces the datasets that are used for examples and experiments throughout the rest of the paper.

Previous works on evaluating \bot performance (discussed in Section \ref{sec:background}) have investigated the use of conversation-level
quality metrics, and individual turn-level assessments of good and bad interactions. In particular, the Interaction Quality (IQ) score 
of \cite{schmitt2015interaction} is explained in this section, and the combined use of ACQIs and IQ score plays a 
major role throughout the rest of the paper.

Section \ref{sec:ACQI_taxonomy} explains the heart of this paper: the design of the ACQI taxonomy.\footnote{The term `taxonomy' is used
here to mean an agreed categorization, and is used for classifications of \bots, intents, and failure states, in addition to the quality indicators and actions
of the ACQI taxonomy. There are not always tree-like `taxonomic' relationships between these categories, though sometimes there are.}
This explains the motivation and decisions behind ACQIs, including how they are made to be actionable, explainable, and to give feedback
that is specifically tailored to the \bot in question. Section \ref{sec:ACQIneedsIQ} describes the annotation work for ACQI datasets,
analysis of the ACQI distributions, and the work on combining ACQI and IQ scoring to distinguish those 
parts of the dialog that need particular improvement.

Section \ref{sec:results} describes experiments on automatically predicting ACQIs based on the annotated datasets and features extracted from the dialog text.
We demonstrate that correct ACQI labels can be predicted with a weighted average f1-score of 79\%, 
and demonstrate the effectiveness of textual features to predict labeled IQ scores (1 to 5) with an average accuracy of 60\%. 
Analyzing the distribution of ACQIs and suggested actions when IQ drops shows that the number of potential improvement strategies to 
evaluate could be reduced by up to 81\%, depending on the accuracy of the ACQI and IQ classifiers. 
We argue that tools built using these approaches could improve the effectiveness and reduce cognitive 
burden on bot-builders. 

\section{Conversational AI Systems and Challenges for Bot-Builders}
\label{sec:conv_ai_systems}

Since the year 2000, \bots (often called chatbots) have gone from mainly research demonstration systems to include various user-facing commercial offerings.
Dialogue systems fall into three broad categories \citep{deriu2020survey}: 
conversational agents, question answering systems, and task-oriented systems (which are the topic of this paper). 
Each category has corresponding dialog quality measurement strategies. 
Conversational agents, which often receive the most attention in news articles when released,
are typically unstructured and open-domain, with no particular objective other than an engaging conversation.
Question answering systems can be evaluated by considering the accuracy of answers given.
Task-oriented systems, which are the focus of this article, typically have a rigid structure and a limited scope. 
They are built to resolve the consumer's issue, answer questions, route to an appropriate representative, or guide the user through a task as efficiently as possible. 
Like question answering systems, task-oriented systems have clear `failure' cases: just as a question answering system can fail to answer a question,
a task-oriented system can fail to complete a task.

Task-oriented systems are prevalent in customer service: ideally, they can automate simple tasks like routing requests to the right agent,
freeing up human customer service specialists to handle more demanding situations. Several technology companies offer \bot services to 
support such chatbots. These include offerings from large and general technology companies such as 
Microsoft LUIS, IBM Watson, Google DialogFlow, and from more specialist providers such as Salesforce, Intercom, and \LP.

While there are several differences between these platforms, there are some typical themes:
\begin{itemize}
    \item The platforms provide various tools and widgets for customers to build and deploy their own \bots.
    \item The dialogs built in this way are `designed' or `scripted'. The process of building a dialog involves declaring
    various steps, inputs the user can be encouraged to make at those steps, and what action the system should take in response.
    \item This process admits the possibility of error or failure states, when then input from a user is something the system
    (knowingly or unknowingly) cannot process successfully.
\end{itemize}

\begin{figure}
    \centering
    \fbox{\includegraphics[width=0.8\linewidth]{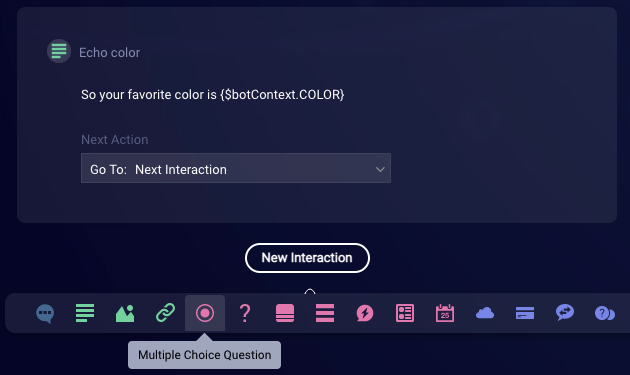}}
    \caption{A bot-building interface where the bot-builder is about to add a multiple choice question}
    \label{fig:conversation_widgets}
\end{figure}

A simple example might be that the dialog is at a stage where the user is asked to pick from a list of available options, such as 
``Enter 1 for English, Ingrese 2 para Español''. In this case, any entry other than the numbers `1' or `2' would be a problem.
This problem can be addressed in a few ways. A traditional keyboard interface might say ``Option unrecognized, please type 1 or 2.''
A form-filling approach might be to use a ``Select from Menu'' element instead of a ``Prompted Text Entry'' element, so that 
the customer can only give input that corresponds to the actions the \bot can take right now. An example from the \LP Conversation Builder
is shown in Figure \ref{fig:conversation_widgets}. The user is about to add an new interaction to 
the conversation so far, has a choice of widgets including those for text entry, scheduling, payments, and various formats for asking questions 
and tracking the responses. There are analogous features in other bot-building platforms.

It is crucial  to note that \bot platforms are typically used by customer support specialists, not the customers themselves.
In the rest of this paper, 
these users will be described using the colloquial but industry-wide term `bot-builders'. This is emphasized because one of the ways
to improve these \bots is {\it to provide tools that make bot-builders more effective}. The work described in this paper shows that
two established approaches to measuring conversations, scoring individual actions and assessing the conversation quality as a whole,
can be combined to provide such a tool that helps bot-builders to identify and fix particular pain-points in a \bot.
The intuitively appealing assumption that useful actionable feedback about \bots can be produced through (semi-) automated methods
is supported by \cite{hockey-etal-2003-targeted}, 
which shows that users of a task-oriented dialog system who received actionable feedback in failure cases outperformed the control group that did not.

\section{Datasets Used in This Work} 
\label{sec:data}

The datasets used in this paper are introduced here and referred to throughout this paper.
The following two datasets were annotated and used in the prediction experiments in subsequent sections.
Statistics about these datasets are summarized in Table \ref{tab:dataset_summary}.

\subsection{\lego}
    
\lego represents a portion  of the CMU Let's Go Public DataSet \citep{Raux2005LetsGP,Raux2006DoingRO}, which was instrumental in the development
of the Interaction Quality (IQ) score of \cite{schmitt2015interaction}, explained in Section \ref{sec:background}.
    
Let's Go Public is a record of phone-mediated 
customer-service interaction between an automated dialogue system and callers drawn from the general population in the vicinity of Pittsburgh, Pennsylvania.
It has been studied intensively since its initial creation, and subsets have been used, in particular, to support analysis and prediction
of IQ \citep{Stoyanchev2017PredictingIQ}. 

The \lego dataset still represents many of the challenges of deploying task-oriented spoken language systems `in the wild':
\begin{itemize}
    \item The callers are drawn from the general population.
    \item The task-at-hand is authentic: callers have a presumed need to access bus information.
    \item The callers are using standard personal or public telephones in real-world settings that include such challenges as third party speech, television programs in the background, very variable audio quality, and irritated and or amused speech.
\end{itemize}

Things that have changed since the creating of \lego include: 
\begin{itemize}
    \item Speech recognition technology is far more capable now than it was when Let's Go began.
    \item The public is much more familiar with conversational AI systems. 
    \item Both speech-mediated and text-mediated dialogue systems are commercially important and widely deployed, so lessons learned from Let's Go have greater potential impact.
\end{itemize}

Previous approaches for modeling quality in \lego have included features automatically extracted from the speech recognition and dialog system \citep{Stoyanchev2017PredictingIQ}.
We have chosen to use only features derived from the text itself from 
a transcribed version of \lego, as this approach is applicable to text-based dialog systems including
those developed by \LP, and avoids using system implementation decisions as features from the dataset itself. 

\subsection{Datasets from \LP \bots}

An approved\footnote{The approval and annotation process includes checking all aspects of legal compliance and scrubbing personally identifiable information (PII) from messages.} 
collection of transcripts of customer-service conversations with \LP \bots was also extracted and annotated. 

We have intentionally chosen a set of bots that serve different functions, come from different industries, and have different overall quality 
(as measured by final IQ score). To this end, 130 conversations were chosen from each of 4 \bots, giving a total of 520 conversations, a comparable
number of conversations to those used from the \lego dataset, though the \LP conversations themselves on average have fewer turns (Table \ref{tab:dataset_summary}). 

These dialog systems often respond with structured content, such as embedded HTML with buttons, toggles, or drop-down menus.  In these cases we represent the text of the options separated by `***'. For example: \textit{``BUTTON OPTIONS *** Main Menu *** Pick a Color *** Pick a different item''}. 

\begin{table}
\begin{footnotesize}
    \caption{Summary description and statistics on parts of CMU \lego and \LP datasets annotated and used in this work}
    \label{tab:dataset_summary}
    \begin{center}
    \begin{tabular}{|c|c|c|c|c|}
        \hline
         \textbf{Bot} & \textbf{Vertical} & \textbf{Bot Functions} & \textbf{Avg \# Turns}& \textbf{\# Conversations}  \\
         \hline\hline
         \lego & Travel & Transactional  & $13.67 \pm 11.25$ & $541$ \\ \hline
          
         Junior Sales Assistant & Retail & Data Gathering, Routing & $4.95 \pm 2.13$ & 130  \\ \hline
         Help and Route & Travel & Routing, FAQ  & $5.89 \pm 3.71$ & 130  \\ \hline
         Router & Tech & Routing & $2.25 \pm 1.45$ & 130 \\ \hline
         Food Expert & Food Industry & Transactional (Ordering) & $5.44 \pm 3.31$ & 130 \\ \hline
        
    \end{tabular}
    \end{center}
\end{footnotesize}
\end{table}

\section{Previous Approaches to Evaluating Dialogue Systems, Including Interaction Quality (IQ)}
\label{sec:background}

Meaningful evaluation of automated dialogue system performance has long been recognized as crucial to progress in dialogue system research
\citet{deriu2020survey} and is an active area of development, including the recent introduction of the `sensibleness and specificity' metric of \cite{adiwardana2020towards}. 
Since the introduction of bot-building platforms as described in Section \ref{sec:conv_ai_systems}, it has also become a daily concern for
customer service bot-builders.

One of the most common industry measures for \bot effectiveness is the {\it automation rate} or {\it containment rate}. Containment rate directly affects the cost savings that a business may make through automation, 
for example ``Over three years and a conservative 25\% containment rate, the
cost savings is worth more than \$13.0 million to the organization \citep{forrester2020watson}.'' However,
analysis of containment or automation rates is still relatively rare in the research literature, and is mainly
found in evaluation of speech / voice driven \bots \citep{pieraccini2009we}. The tradeoff between
the automation rate and the number of setbacks a user can encounter is analysed in \citet{witt2011global} ---
also in speech. The relative lack of research literature on containment rates can be easily attributed to different
settings and incentives. The case study presented in \cite{forrester2020watson} notes that
{\it ``Over three years and a conservative 25\% containment rate, the
cost savings is worth more than \$13.0 million to the organization.''} By contrast, most \bots in research settings 
do not have human agents to respond to escalations, so the systems cannot escalate, and containment cannot be measured.

For the task-oriented dialogue systems that are our primary concern, there is an underlying concept of giving a `right' response.
Early evaluation work based on this idea compared the actual responses to a predefined key of reference answers \citep{hirschman-etal-1990-beyond}. 
The portion of matches to the key gave the measure of performance. Well known weaknesses of this approach include being specific to particular systems, domains and dialogue strategies.
More portable strategies that measure inappropriate utterance ratio, turn correction ratio, or implicit recovery \citep{Danieli_Gerbino1995, hirshman_pao1993, polifroni1992, shriberg1992} are intellectual ancestors of the ACQI part of our approach, in that they identify events that are indicators of the quality of the conversation.  
Both of these early approaches share the limitation of being unable to model or compare the contribution that the various factors have on performance. 

The PARADISE approach \citep{paradise} overcomes this limitation by using a decision-theoretic framework to specify the relative contribution of various factors to a dialogue system's  overall performance. This and other ideas introduced by PARADISE, such as
separating the accomplishment of a task from how the system does it, 
support evaluation that is portable across different systems, domains and dialogue strategies.
We have tried to emulate some of these best practices by defining ACQIs that avoid being too implementation-specific,
though without insisting that all ACQIs must be relevant to all \bots (see Section \ref{sec:ACQI_taxonomy}). For example,
in contrast with \cite{Stoyanchev2017PredictingIQ},
we have chosen not to utilize any features extracted from the dialog systems themselves (e.g. failed intent match from the 
natural language understanding system) and rely solely on the text of the dialog. 

\subsection{Interaction Quality (IQ) Score}
\label{sec:iq_score}

More recent work has developed several dialog quality measurement strategies which are categorized by \cite{bodigutla2019multi} as: 
sentiment detection; per turn dialog quality (e.g. Interaction Quality \citep{schmitt2012parameterized} and Response Quality (RQ));
explicitly soliciting feedback from the user; task success; and dialog-level satisfaction ratings as in PARADISE \citep{paradise}. 
These methods are useful but have well-known limitations: for example, sentiment analysis on messages misses many problems, 
there is response bias in user polling, and negative outcomes weigh more in the consumer's mind than positive ones \citep{han2020customer}.

The method in this paper most directly builds on the IQ method of \cite{schmitt2012parameterized,stoyanchev2019predicting}. 
The methods involve annotating a conversation, typically adding one point for a good interaction and subtracting
a point for a bad interaction, with some exceptional cases, for example when the dialog `obviously collapses'.
A benefit of these methods is that they can help to identify {\it where} there are problems in a \bot:

\begin{quoting}
While the intention of PARADISE is to evaluate and compare
SDS or different system versions among each other,
it is not suited to evaluate a spoken dialogue at arbitrary points during an interaction. \citep{schmitt2012parameterized}
\end{quoting}

Our work can be seen as an extension of the IQ approach \citep{schmitt2011modeling} in our use of a running dialogue quality measure. 
ACQI improves over IQ by recommending how to resolve a problem instead of only identifying where it exists.

\section{The Design of the ACQI Taxonomy}
\label{sec:ACQI_taxonomy}

This central section describes the design of the ACQI taxonomy and how it improves over the IQ system introduced above.

A running score like IQ allows bot-builders to identify dialog system responses where there is a meaningful decrease in score, but does not 
provide direct diagnosis of the problems that may be there. If nothing other than the running score is available, 
bot-builders have little option other than to manually review, form their own taxonomy of failure causes, and come up with appropriate fixes. This process in practice typically takes days or weeks to complete, and is prioritized largely by operator intuition.

To make the problems explicit and to suggest solutions, we introduce {\it Actionable Conversation Quality Indicators} or ACQIs.
ACQIs highlight moments in chatbot conversations that impact customer experience. Our ACQI taxonomy can be found in Table \ref{tab:acqi_taxonomy}. 
Note that for each ACQI, there is an associated set of an actions a bot-builder can take to mitigate the issue.
The ACQIs and the taxonomy dimensions are all derived from analyzing the user experiences directly. As well as making them
actionable, this is motivated by wanting to make the issues aggregable, so that bot-builders can analyze aggregated statistics about
conversations and prioritize fixing the most prevalent issues accordingly. By exposing predicted ACQIs in an appropriately aggregated format, we empower 
bot-builders to make more data-driven decisions when improving their \bots. 
Not all ACQIs are relevant to all \bots: for example, the CMU \lego system does not support transfer to a human agent, so `Bad Transfer' cannot occur
(though `Set Transfer Expectations' can still potentially be useful, even to say that there are no human agents available).

\begin{table*}[ht]
\begin{center}
    \caption{ACQIs with their associated actions in example text and spoken \bots}
    \label{tab:acqi_taxonomy}
\begin{adjustbox}{width=\textwidth}
\begin{footnotesize}
\begin{tabular}{|>{\raggedright}p{1.5cm}
                |>{\raggedright\arraybackslash}p{4.8cm}
                |>{\raggedright\arraybackslash}p{4.8cm}
                |>{\raggedright\arraybackslash}p{4.8cm}|}
                
        \hline
        \textbf{ACQI} & \textbf{Description} & \textbf{\LP Action}& \textbf{CMU \lego Action} \\
        
        \hline\hline
        Does Not Understand & 

        The bot says that it does not understand the consumer's response and says something like `I didn't get that' or `I don't understand'. This also includes when the bot's response is incorrect and it is clear that the bot has misunderstood what the consumer said. &
        Add/Remove Features, Update NLU Taxonomy, Update NLU Training Data, Add/Remove/Update Confirm Consumer Statement(s), Add FAQs, Add/Update Bot Statements, Add/Update Bot Prompts, Add/Update Max No Match Behavior & 
        Add/Remove Features, Update ASR Training Data, Update NLU Taxonomy, Update NLU Training Data, Add/Remove/Update Confirm Consumer Statement(s), Add/Update Consumer-Requested Repeat, Add FAQs, Add/Update Bot Statements, Add/Update Bot Prompts, Add/Update Max No Match Behavior \\ 
        
        \hline
        Input Rejected  &  
        The bot does not accept a consumer response to menu options, forms, or other structured content, such as `1, 2, 3, 4' , `a, b, c, d', or `yes, no`. &
        Improve Response Flexibility, Add/Update Bot Statements, Add/Update Bot Prompts, Add/Update Max No Match Behavior &
        Improve Response Flexibility, Add/Update Bot Statements, Add/Update Bot Prompts, Add/Update Max No Match Behavior \\
        
        \hline
        Ignored Consumer  & 
        The dialog system does not recognize a free text consumer response and asks an unrelated question. &
        Set Delay Expectations, Enable/Disable Context Switching, Add FaaS code, Confirm Successful Bot Handoff, Handle Backend Errors &
        Set Delay Expectations, Enable/Disable Context Switching, Add/Update Turn-Taking Signals, Add FaaS code, Add/Update Max No Input Behavior, Handle Backend Errors\\ 
        
        \hline
        Restart  & 
        When explicitly asked for by the consumer and only used when the conversation starts over within the same engagement with the bot. Includes when the bot goes back to the main menu. &
        Add/Remove/Update Confirm Consumer Statement(s), Add/Update Conversational Navigation, Add/Update Help, Add/Update Max Restart Behavior &
        Add/Remove/Update Confirm Consumer Statement(s), Add/Update Conversational Navigation, Add/Update Help, Add/Update Max Restart Behavior\\ 
        
        \hline
        Bad Transfer  & 
        The dialog system attempts to transfer the consumer to an agent, but either leaves them hanging or abruptly ends the chat. It might also fail to tell them EARLY enough in the conversation that there are no agents available at that hour. &
        Set Transfer Expectations, Confirm Successful Transfer &
        Transfer is not possible in the CMU system, since its only use is out of hours. 
        Set Transfer Expectations makes sense, but the issue never arose, because Bad Transfer was never used.
        \\ 

        \hline
        Unable to Resolve &
        The bot explicitly states that it cannot provide the information the consumer requested and does not offer to transfer to an agent. &
        Add/Remove Features, Offer External Solutions, Set Transfer Expectations &
        Add/Remove Features, Offer External Solutions, Set Transfer Expectations \\ 
        
        \hline
        Ask for Information &
        The bot asks the consumer for information. &
        Remove Unnecessary Questions, Set Delay Expectations, Enable/Disable Context Switching, Add FaaS code, Confirm Successful Bot Handoff, Handle Backend Errors &
        Remove Unnecessary Questions, Set Delay Expectations, Enable/Disable Context Switching, Add/Update Turn-Taking Signals, Add FaaS code, Add/Update Max No Input Behavior, Handle Backend Errors \\ 
        
        \hline
        Provide Assistance & 
        Bot responds to consumers query, attempting to fulfill  their intent. & 
        Add/Remove/Update Validate Bot Resolution, Add/Remove Features, Update NLU Taxonomy, Update NLU Training Data, Add/Remove/Update Confirm Consumer Statement(s), Add FAQs, Add/Update Bot Statements, Add/Update Bot Prompts & 
        Add/Remove/Update Validate Bot Resolution, Add/Remove Features, Update ASR Training Data, Update NLU Taxonomy, Update NLU Training Data, Add/Remove/Update Confirm Consumer Statement(s), Add FAQs, Add/Update Bot Statements, Add/Update Bot Prompts \\ 
        
        \hline
        Ask for Confirmation & 
        Bot asks for user confirmation of input. Includes when the bot asks if the information provided is correct. & 
        Update NLU Taxonomy, Update NLU Training Data, Add/Remove/Update Confirm Consumer Statement(s), Add/Update Bot Statements, Add/Update Bot Prompts & 
        Update ASR Training Data, Update NLU Taxonomy, Update NLU Training Data, Add/Remove/Update Confirm Consumer Statement(s), Add/Update Bot Statements, Add/Update Bot Prompts \\ 
        \hline
\end{tabular}
\end{footnotesize}
\end{adjustbox}
\end{center}
\end{table*}



The ACQI taxonomy is inspired from a variety of sources, including literature on the (human) evaluation of open-domain dialog systems, consultation with experts, and ongoing feedback from expert users involved in bot-building.

We carried out a series of interviews with bot-builders and assembled a repertoire of tuning actions based on their practices. 
We identified 28 distinct actions bot-builders take at \LP. In the case of \lego, we consulted with 2 domain experts and identified 31 actions.
For each of our ACQIs we assigned a set of actions that the bot-builder may make (Table \ref{tab:acqi_taxonomy}). 
For each of our ACQIs we assigned a set of actions that the bot builder may take (Table \ref{tab:acqi_taxonomy}). The mapping showed that our ACQIs could guide bot builders to take 23 of 28 possible actions for \LP systems, and 25 of 31 possible actions for CMU.

The mapping from ACQIs to actions was created by a \LP expert in improving dialog systems, and later refined with further feedback from \LP bot-builders. 
Because IQ+ACQI will be aggregated and used to improve consumer experience (CX) we put an explicit emphasis on CX and actionability. 

\cite{finch2020towards} list 21 dimensions across 20 publications that human evaluators have used to measure the quality of open-domain dialog systems.
While these dimensions are not used directly, they partially inspire our taxonomy of failure states (Table \ref{sec:ACQI_taxonomy}, leftmost column). 
For instance the `Doesn’t Understand' failure state is inspired by Coherence \citep{luo2018auto,wu2019proactive}, Correctness \citep{liu2018knowledge,wang2020improving}, Relevance \citep{moghe2018towards,lin2019moel,qiu2019training}, Logic \citep{li2018syntactically} and Sensibleness \citep{adiwardana2020towards}.
Other elements of our taxonomy are informed by \cite{jain2018evaluating} which provides a set of best practices when developing dialog systems for messaging based bots. We separated `misunderstanding' into `Input Rejected', `Ignores Consumer' and `Does Not Understand' categories, because each of these require different 
actions that can mitigate the understanding issue. Such separation of failure states highlights the design-for-actionability of the ACQI taxonomy.


\subsection{Desirable Properties of a ACQI Taxonomy}

A good ACQI taxonomy should be actionable, easy to understand, have highly bot-dependent ACQI incidence rates, and have a significant impact on consumer experience. In the following sections we will justify these properties, and provide measurement strategies where appropriate. For some ACQIs and dialog systems it may be appropriate to bypass the need for annotation by automatically extracting ACQIs from the system logs. An example for this would be an ACQI that indicates that the NLU returned a confidence score less than a predefined threshold and responded with a request for the user to rephrase their intent. 

\subsubsection{Actionable}

Without actionable ACQIs, bot-builders are left to the time consuming process of reviewing large amounts of transcripts and/or a careful analysis of the model's feature space, from which they can attempt to deduce what mitigation strategies are appropriate. Unfortunately many features are not actionable from the bot-builder's perspective. For instance, conversation length can be highly predictive of conversation quality, but from their perspective, there are no clear steps to uniformly reduce conversation length. The problem is that conversation length is not the cause of a bad conversation, it is a consequence of it. 

For example, for the dialog system `Food Expert', the initial turn was predicted to have ignored a consumer's initial intent. 
This can lead to longer conversations, as can an order with complicated modifications, but the associated dialog quality improvement 
strategies for the operator are quite different. ACQI allows for separation and sizing of these situations: the second is varied and complicated,
whereas the first is a single problem with a negative affect on IQ.
From the bot-builder's perspective this one is a relatively easy fix: they just need to make sure that intent recognition is applied to the first customer utterance. 


\subsubsection{Easy to understand} 

For bot-builders, understanding is a necessary condition for fixing an issue. The terms used in the ACQI taxonomy are
deliberately chosen to be familiar to bot-builders, and have been refined with use to add clarity where requested.
The relative success of this effort is reflected in the encouraging inter-annotator agreements reported in 
Table \ref{tab:annotator_agreement} below. However, this finding is potentially influenced partly by the use of
experts in bot-building and conversation modeling as annotators. This has yet to be corroborated with less experienced bot-builders.


\subsubsection{Bot Dependent ACQI Rates} 

\begin{figure}[t]
\centering
\includegraphics[width=0.4\linewidth]{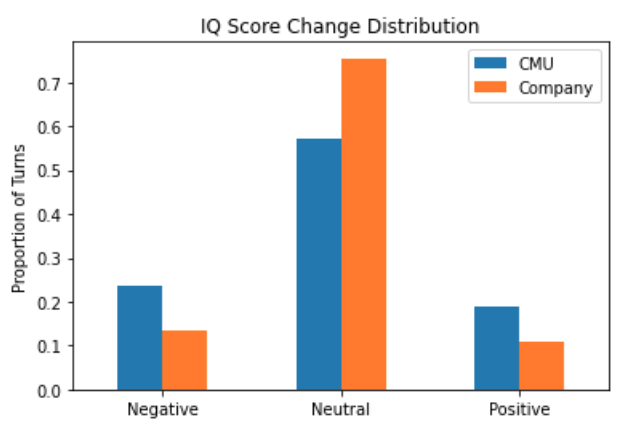}
\vspace{0.2in}
\caption{Distribution of negative/neutral/positive score changes grouped by CMU and \LP dialog systems.}
\label{fig:ACQI_frequency}
\end{figure}

As the purpose of the ACQI taxonomy is to guide bot-builders to appropriate fixes it is imperative that the rates at which ACQIs are present are bot-dependent. From Figure \ref{fig:ACQI_frequency} we can see this is indeed the case. As our ACQI taxonomy is meant to fit a wide breadth of task-driven dialog systems, teams working on very specific \bots may wish to use a subset, or create their own smaller taxonomy. 
For instance, if the dialog system does not allow for transfers (this is the case for \lego) `Bad Transfer' should be excluded from the taxonomy.

\subsubsection{Impact on Customer Experience} Each ACQI should have an actionable relevance to CX. Figure \ref{fig:score_change_dist} illustrates this property for our ACQI taxonomy as each score change distribution differs from the overall distribution (topmost bar). Several elements of the taxonomy only indicate a poor customer experience in particular circumstances. For example, the `Ask for Confirmation' is more likely to indicate a negative CX when it occurs multiple times within a given dialog (Figure \ref{fig:confirmation_impact}). 

This concludes the summary of the ACQI taxonomy itself. In practice we found that ACQIs were most
reliably predicted and effectively used in combination with a running quality score. This work is described next.

\section{Combining ACQIs with IQ in Conversational Datasets} \label{sec:ACQIneedsIQ}

The section describes the work done on annotating ACQIs along with IQ, which turned out to be necessary for
distinguishing those ACQIs that warrant action. This is due to the fact that dialog context matters. For instance, a system attempting to correct a misunderstanding of a malformed user statements is quite different than the system failing to understand an unambiguous answer to a direct question.

\todo{Can we add anything concrete about annotating ACQIs alone?}

\todo{Was the \lego dataset added / included as part of adding IQ to the mix, or did we start with both \lego and \LP datasets?}

\subsection{Observations from Annotating ACQIs}

Our initial approach to building a model for finding ACQIs and making associated recommendations to bot-builders was based on the
assumption that particular ACQIs are bad for the user experience and should be avoided. This turned out not to be the case. As the conversations 
from the \LP datasets were annotated, we observed that ACQIs alone may be bad or good or neither. 
For example, `Ask for Confirmation' is good when the user input is genuinely vague,
but is bad when the user input was clear and the system should have understood. For example, asking the user to confirm that 
`next Wednesday' refers to a particular calendar date is sometimes helpful (especially if today is Tuesday!). 
Asking if the user's
response was `3' if the user just selected `3' can by contrast be obvious and irritating.
If a consumer uses `Restart' one time, 
it can be seen as a good signal that the consumer requested to go back and the system responded appropriately,
but if it is used more frequently it can be a signal that their request is not being properly handled.
Based on ACQI designations alone, bot-builders cannot always 
be sure whether a corrective action is needed. Essentially, ACQIs are context dependent and combine (with context and themselves) nonlinearly. In this work we explore the context dependency via the relationship with IQ. We leave a more structured statistical modeling framework for future work.

\subsection{Annotating ACQIs and IQ Together}

To mitigate the issue of ACQI instances not being universally good, bad, or neutral, it was decided to combine ACQIs with an overall
quality score such as Response Quality \citep{bodigutla2020joint} or Interaction Quality \citep{schmitt2015interaction}, discussed in
Section \ref{sec:background}. For these purposes, IQ was chosen, partly because of the availability of the \lego dataset
and comparable prior work. 

\begin{figure}
    \centering
    \includegraphics[width=0.5\linewidth]{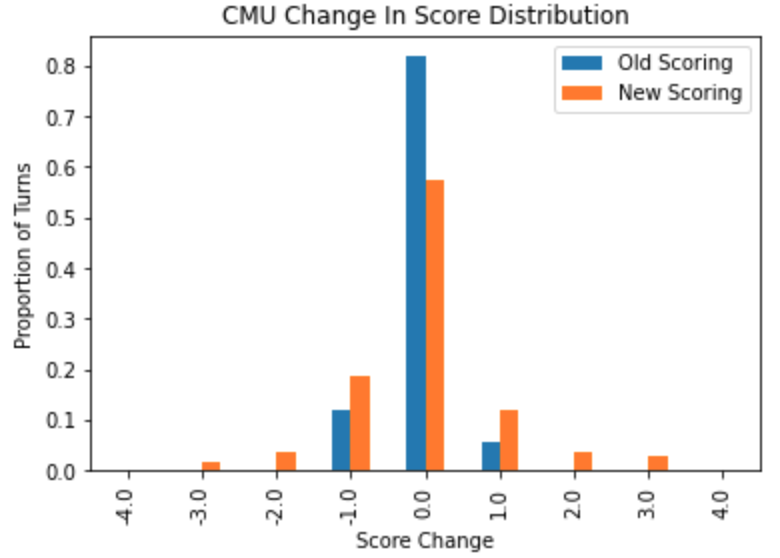}
    \vspace{0.25in}
    \caption{Change in IQ score distribution between IQ annotations in \cite{LEGOv2} and the work reported in this paper.}
    \label{fig:iq_score_change_dist}
\end{figure}

There are several differences between the guidelines given in \cite{schmitt2011modeling} and our own. \todo{See the transcribed \lego resource paper for a full discussion add Chris resource paper.} Most importantly, the removal of all guidelines relating to how much the score can be increased/decreased. Due to their restrictive guidelines the change in IQ is almost entirely (99.7\%) in increments of 0,1, or -1. From observations in negativity bias \citep{negativity_bias} we know this does not reflect how consumers feel when encountering undesirable behaviours. The impact of the removal of these guidelines can be seen in figure \ref{fig:iq_score_change_dist}. In spite of these changes our annotator agreement has slightly increased with $\rho=.69,.72$ for them and $\rho=.76$ for us.

We also altered the `dissatisfaction scale' used by \cite{schmitt2011modeling} (5-satisfactory, 4-slightly unsatisfactory, 3-unsatisfactory, 2-very unsatisfactory, 1-extremely unsatisfactory ) to 5-good, 4-satisfactory, 3-bad, 2-very bad, 1-terrible. We added the `good` category as we want our IQ score to show that there is {\it room} for improvement even if the recommended improvement is not currently technically feasible.
The guideline dictating that each conversation should start with a satisfactory rating was removed as our dialog systems the consumer is often the party that initiates the dialog (in \lego the system always opens the dialog with a greeting), so the \bot can make a mistake with interpreting the very first message in the conversation.

\subsection{Annotation and Inter-Annotator Agreement}

Annotation of the 531 \lego conversations and 520 \LP conversations was carried out by three annotators employed by \LP who are experienced live customer service agents.
Following \cite{schmitt2011modeling} we took the median score of our 3 annotators as ground truth. 
For 3 way ties for ACQI (6.4\% of turns) we chose to use the most common label for that particular \bot out of the labels the annotators have chosen.
So as in \cite{schmitt2011modeling} and \cite{stoyanchev2019predicting} we use third parties rather than 
the users of the \bot to judge the turn by turn quality, and like \cite{stoyanchev2019predicting} our annotators are expert rather than crowdsourced. 
The biggest differences between their annotation work and ours is that the guidelines for the running IQ score are simplified, 
and we include an additional annotation task for ACQI to recommend an appropriate fix.
The averages of the minimum and final IQ scores are shows in Table \ref{tab:iq_results_summary}, which shows that the CMU \lego has the lowest
dips in IQ score during a conversation, but by the end of the conversations, the IQ for the various bots is quite similar.

\begin{table}
\centering 
\caption{Average Minimum and Final IQ Scores from Annotation}
\label{tab:iq_results_summary}
\begin{tabular}{|c|c|c|}
    \hline
     \textbf{Bot} & \textbf{Ave Final IQ Score} & \textbf{Ave Min IQ Score}  \\
     \hline\hline
     \lego & $3.78$ & $2.44$\\ \hline
      
     Junior Sales Assistant & 3.43  & 3.15  \\ \hline
     Help and Route & 3.79 & 3.44   \\ \hline
     Router & 3.74 & 3.61 \\ \hline
     Food Expert & 4.08 & 3.48 \\ \hline
    
\end{tabular}
\end{table}

In spite of having similar
overall IQ outcomes, the CMU \lego \bot has fewer neutral neutral steps, and more positive and negative turns, whereas \LP \bots have more neutral turns.

We measured inter-annotator agreement to measure the clarity of our ACQI taxonomy and IQ rules when applied to real \bots.
The results have a Cohen's  Kappa (CK) of .68, which is substantial agreement according to \cite{landis1977measurement}. 
See Table \ref{tab:annotator_agreement} for more details. 

\begin{table}[t] 
  \centering
    \caption{Annotator Agreement for annotating IQ and ACQI, showing linear weighted cohen kappa (LWCK), unweighted average recall (UAR), spearman rank correlation ($\rho$) for IQ and cohen kappa (CK) for ACQI. Following \citet{schmitt2015interaction} we take the average agreement across each pair of annotators.}
    \label{tab:annotator_agreement}
    \begin{footnotesize}

    \begin{tabularx}{\linewidth}{|X|c|c|c|c|c|X|}
        \hline
          & \multicolumn{3}{|c|}{IQ} & {ACQIs} \\
         \hline
         \textbf{Bot} & \textbf{UAR} & \textbf{LWCK} & $\rho$ & \textbf{CK}  \\ 
         \hline\hline
         CMU  & .61 & .65 & .76 & .71 \\ \hline
         Junior Sales Assistant & .46 & .34 & .21 & .64    \\ \hline
         Help and Route & .55 & .50 & .53 & .67  \\ \hline
         Router & .39 & .49 & .64 & .78  \\ \hline
         Food Expert & .60 & .63 & .71 & .60 \\ \hline \hline
         All (macro-averaged) & .52  & .52 & .57 & .68 \\ \hline
    \end{tabularx}
    \end{footnotesize}
\end{table}

\subsection{Analysis of Combined ACQI and IQ Annotations}

We are now in a position to analyze the correspondence between individual ACQIs and changes in the overall IQ score.
By looking at the turn to turn change in IQ given the presence of a particular ACQI we gain a more nuanced understanding of how a dialog system is performing. In previous attempts to measure dialog quality a lack of understanding and an inability to resolve a consumers issue (task completion) are taken to be unequivocally bad. This is not always the case. 

\begin{figure*}[t!]
\centering
\vspace{0.2in}
\includegraphics[width=0.9\linewidth]{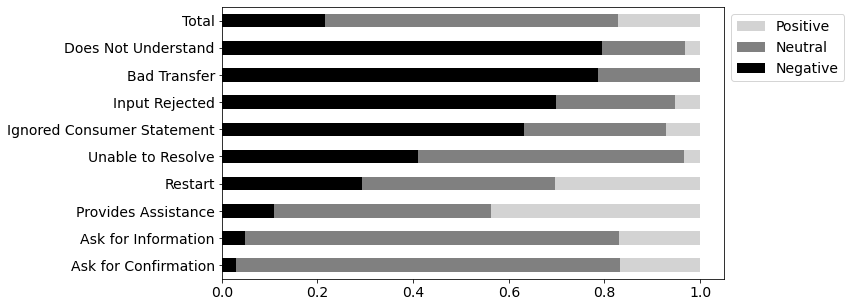}
\caption{ACQIs from Table \ref{tab:acqi_taxonomy} along with the proportions of each that were aligned with positive, negative, and neutral
changes in IQ. Note that for the above graphic we excluded any turn whose preceding score IQ was a 1 or 5.}
\label{fig:score_change_dist}
\end{figure*}

From Figure \ref{fig:score_change_dist} we see that with the exception of `Bad Transfer' (which was not annotated as negative) 
all of our ACQIs can be positive, negative or neutral. 
`Does Not Understand', `Ignored Consumer Statement' and `Input Rejected' are usually negative, but can be reasonable responses
if the consumer is typing things that are out of scope or incomprehensible.
However, they are negative when the consumers utterance should have been understood by the system. 
`Unable to Resolve' is positive when the system correctly identifies that the consumer's request is out of scope.
`Provides Assistance' is positive if the assistance was requested and appropriate, but negative if it was not, or if the assistance has already been provided. 
Similarly, `Ask for Information' can be positive or negative, depending on the relevance of the requested information.

Analyzing the annotated datasets by ACQI and \bot is also instructive (Table \ref{tab:dist_ACQI_given_score_decrease}). For most of the \bots,
`Does Not Understand' is the largest category of ACQI associated to a decrease in score. The {\it Junior Sales Assistant} is 
an exception, and for this, 54\% of the ACQIs are
marked as `Provides Assistance', indicating that this bot is making too many improper transfers or inappropriate suggestions to the customer.
While having the score alone would be somewhat valuable in locating this issue, 
the ACQI gives additional guidance on what steps can be taken to mitigate the undesirable behavior.

\begin{table*}[t]
  \caption{Distribution of ACQIs given a decrease in IQ score.}
    \label{tab:dist_ACQI_given_score_decrease}
    \begin{center}
    \begin{footnotesize}
    \begin{tabularx}{0.8\linewidth}{|p{3.5cm}|X|X|X|X|X|}
        \hline
         \textbf{Dialog System} & \textbf{CMU \lego} & \textbf{Junior Sales Assistant}& \textbf{Help and Route}& \textbf{Router}& \textbf{Food Expert} \\ 
         \hline\hline
         Does Not Understand & 0.655 &	0.196 &	0.631 &	0.667 &	0.557 \\ \hline
         Ignored Consumer Statement	 & 0.093 &	0.009&	0.046&	0.000&	0.076   \\ \hline
         Ask for Information & 0.078 &	0.140&	0.077&	0.000&	0.165  \\ \hline
         Input Rejected & 0.057&	0.009&	0.000&	0.000&	0.013 \\ \hline
         Unable to Resolve & 0.042&	0.000&	0.000&	0.000&	0.051 \\ \hline
         Ask for Confirmation & 0.034&	0.000&	0.000&	0.000&	0.089  \\ \hline
         Provides Assistance & 0.024	& 0.542& 	0.246&	0.333&	0.051  \\ \hline
         Restart  & 0.017&	0.000&	0.000&	0.000&	0.000   \\ \hline
         Bad Transfer  & 0.000&	0.103&	0.000&	0.000&	0.000  \\ \hline
    \end{tabularx}
\end{footnotesize}
\end{center}
\end{table*}

\begin{figure}[t]
\begin{center}
\includegraphics[width=0.5\linewidth]{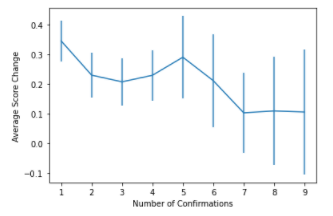} 
\vspace{0.2in}
\caption{Dependence of score change on number of confirmations.}
\label{fig:confirmation_impact}
\end{center}
\end{figure}

We can also analyze combinations of ACQIs and their affect on a conversation. For example, Figure \ref{fig:confirmation_impact} shows
that asking for confirmation once normally leads to an increase in IQ score, but after this the effectiveness decreases and asking for
confirmation more that 6 times can be actively harmful.
It is important to realize that the ACQIs combine nonlinearly. That is, a single instance of an ACQI in a conversation may be healthy,
but multiple copies of it may be a negative indicator. The extent to which the ACQIs have known meaning is the extent to which structural statistical 
models can be built, allowing bot creators the ability to test and refine explicit hypotheses about how the conversational events actually 
aggregate to the \bot user experience. The careful definition and annotation work described so far enables us to start quantifying such effects 
in ways that were not hitherto available. Analyzing such combinations of causes and effects in \bots will be extended in future work.

\section{Experimental Validation of the Predictive Ability of the ACQI and IQ Model}
\label{sec:results}

While the annotation and analysis so far was already able to provide some useful insights, the larger goal of these
efforts is to build systems that can automatically highlight crucial ACQIs in a \bot so that they can be fixed.

There is prior work in predicting IQ from labelled conversations, using a variety of methods.
The original work of \cite{schmitt2011modeling} used Support Vector Machines, and more recent
work has used stateful neural network models including LSTMs \citep{rach2017interaction} and biLSTMs \citep{bodigutla2020joint}. 

\subsection{Vectorized Features from Conversations}

In common with these approaches, we adopted a vector representation for features. However, we deliberately restricted our
model to use only features extracted directly from the conversation text and the annotation, so that the method could be more universally applicable,
and in particular, to be able to use the same annotation setup and featurization processes for data from the \lego and \LP \bots.

For our `text' features (Table \ref{tab:iq_model_performance}), we use the pretrained contextual sentence embeddings of BERT \citep{reimers2019sentence}. Our embedding dimension is 768 for each speaker's response. When predicting ACQI and IQ we concatenate the embeddings for both consumer and dialog system for this turn and the previous. This results in a $4\times 768 = 3072$ dimensional vector. When the previous utterance is unavailable, we use the zero vector 
of the appropriate dimension. This feature vector was used to represent a system where the model uses only surface textual features for the previous two turns per user.

To compare with a system that also uses its observations and predictions of the rest of the conversation so far, we
experimented with adding features derived from the ACQI labels.
For the features `only-gold-acqi' and `only-pred-acqi' we use cumulative counts on the one-hot encoding for the presence of ACQIs. 

These feature sets were also combined as `gold-acqi+text' and `pred-acqi+text',
for which we concatenated the cumulative counts to the contextual sentence embeddings with our cumulative ACQI counts. 

\subsection{Model Training and Prediction}

To get a good representation of many ACQIs across the training and test sets, we used the {\it multi-label Scikit Learn} python package and methodology of
\cite{szymanski2017network}, which supports balanced multi-label train/test splits which we adapted for nested cross-validation \citep{krstajic2014cross}. Utilizing this package we implemented nested cross-validation using 5 cross-validated folds for the inner and outer loop. For predicting ACQI we tested logistic regression, random forest, and xgboost. For IQ we tested the above and a linear regressor. 
We found that the best performing text-based model for predicting both ACQIs and IQ was logistic regression with C (inverse regularization strength) set to 
0.01 and with the `balanced' class weight setting in Scikit Learn.

\subsection{Predicting IQ: Results and Discussion}

For the IQ prediction experiment, results for each of the \bots using the features above are presented in Table \ref{tab:iq_model_performance}.
The results found can also be compared with the annotator agreement findings of Table \ref{tab:annotator_agreement}.

Points to highlight include:
\begin{itemize}
    \item The text-only prediction, which is the easiest to implement, achieves linear weighted Cohen Kappa performance slightly lower but 
    comparable to annotator agreement (average 0.49 vs 0.52).  
    \item Using only ACQI information (either gold or predicted) leads to a loss in recall in all but one \bot, and on average reduces recall by around 10\% using only gold labels and 20\% using only predicted labels.
    \item Though the use of gold labels along with text features leads to improved correlation with annotators (for example, an average 6\% improvement in kappa score), these expectations do not transfer reliably to the more realistic case of
    using predicted ACQI labels as features.
    \item The increase in average recall from using gold labels is small (only 2\%), and the use of predicted labels causes
    average recall to drop by 7\%.
\end{itemize}

These experiments leave much room for optimization and improvement of various kinds, including trying different 
text featurizers, and the number of turns and relative weights of messages used in the vector encoding. 
The important findings are that we can predict the exact IQ score approximately 60\% of the time, and that the use 
of vector embeddings derived directly from the message texts is the most reliable practical method tested for
building features.

\begin{table*}[!ht]
\caption{IQ Model Performance: linear weighted Cohen kappa (LWCK), unweighted average recall (UAR), Spearman rank correlation ($\rho$) for IQ. Model Selection and Hyper-parameter selection was accomplished by nested cross-validation (5 folds).}
\label{tab:iq_model_performance}

\begin{footnotesize}
\begin{center}
\begin{tabular}{|l|l|l|l|l|}
\hline
  \textbf{Dialog System} &        \textbf{Features Used} & \textbf{$\rho$} &   \textbf{UAR} &  \textbf{LWCK} \\ \hline

\multirow{ 5}{*}{CMU} & text &    0.572 & 0.585  &  0.47 \\ \cline{2-5}
 & only-gold-acqi &    0.469 & 0.499 &    0.404 \\ \cline{2-5}
 & only-pred-acqi &    0.296 & 0.396 &    0.242 \\ \cline{2-5}
 & gold-acqi+text &    0.622 & 0.597 &    0.529 \\ \cline{2-5}
 & pred-acqi+text &    0.593 & 0.516 &    0.475 \\ \hline

\multirow{ 5}{*}{Junior Sales} & text &    0.721 & 0.567 &  0.678 \\ \cline{2-5}
 & only-gold-acqi &    0.147 & 0.429 &    0.349 \\ \cline{2-5}
\multirow{ 3}{*}{Assistant} & only-pred-acqi &    0.529 & 0.399 &     0.494 \\ \cline{2-5}
 & gold-acqi+text &     0.84 & 0.599 &     0.797 \\ \cline{2-5}
 & pred-acqi+text &    0.831 & 0.573 &    0.798 \\ \hline

\multirow{ 5}{*}{Help } &  text &    0.569 & 0.503 &    0.495 \\ \cline{2-5}
& only-gold-acqi &     0.25 & 0.363 &    0.285 \\ \cline{2-5}
\multirow{ 3}{*}{and Route}& only-pred-acqi &    0.288 & 0.384 &    0.256 \\ \cline{2-5}
 & gold-acqi+text &    0.673 & 0.553 &   0.597 \\  \cline{2-5}
  & pred-acqi+text &    0.682 & 0.434 &    0.583 \\ \hline

 \multirow{ 5}{*}{Router} & text &    0.621 & 0.346 &  0.585 \\ \cline{2-5}
 & only-gold-acqi &    0.608 & 0.482 &   0.521 \\ \cline{2-5}
 & only-pred-acqi &    0.636 & 0.341 &   0.592 \\ \cline{2-5}
 & gold-acqi+text &    0.765 & 0.635 &   0.679 \\ \cline{2-5}
 & pred-acqi+text &    0.841 & 0.374 &   0.791 \\ \hline

 \multirow{ 5}{*}{Food Expert} &  text &    0.698 & 0.603 &   0.609 \\ \cline{2-5}
 & only-gold-acqi &    0.574 & 0.469 & 0.493 \\ \cline{2-5}
 & only-pred-acqi &    0.514 & 0.435 &  0.448 \\ \cline{2-5}
 & gold-acqi+text &    0.765 & 0.635 &  0.679 \\ \cline{2-5}
 & pred-acqi+text &    0.787 &  0.55 &  0.691 \\ \hline
\hline
 \multirow{ 5}{*}{All Bots} & text &    0.595 & 0.599 &  0.499 \\ \cline{2-5}
  & only-gold-acqi &    0.462 & 0.511 &  0.417 \\ \cline{2-5}
  & only-pred-acqi &    0.326 & 0.408 &  0.274 \\ \cline{2-5}
  &      gold-acqi+text &    0.651 & 0.612 &  0.561 \\ \cline{2-5}
  &      pred-acqi+text &    0.624 & 0.531 &  0.514 \\ \hline
\end{tabular}
\end{center}
\end{footnotesize}
\end{table*}

\subsection{Predicting ACQI: Results and Discussion}

For the IQ prediction experiment, results for each of the \bots using only text-derived features and 
logistic regression as a classifier (as described above) are presented in Table \ref{tab:iq_model_performance}.

Points to note include:

\begin{itemize}
    \item The overall weighted average f1-score is 0.790. We are predicting the correct ACQI nearly 80\% of the time overall.
    \item The accuracy of the classification depends significantly on the support of the class: common ACQIs are predicted much more accurately 
    that rare ones.
    \item Because of this, the macro average performance (not weighted by support) is worse, with an f1-score of just 0.574.
    \item The score can be cross-referenced against Figure \ref{fig:score_change_dist} and Table \ref{tab:dist_ACQI_given_score_decrease} to guide improvement efforts. For example, `Does Not Understand' occurs relatively frequently and with overwhelmingly negative impact on IQ score, but the F1-score for predicting this
    class is only 0.509. Improving ACQI classification performance for this class would therefore be especially impactful.
\end{itemize}

\begin{table*}[!ht]
\begin{center}
\caption{ACQI Model Performance}
\label{tab:cqi_model_performance}
\begin{footnotesize}
\begin{tabular}{|p{2.4cm}|r|r|r|r|}
\hline
                      \textbf{ACQI} &  \textbf{Precision} &  \textbf{Recall} &  \textbf{F1-score} &  \textbf{Support} \\ \hline
       Ask for Information &      0.948 &   0.896 &     0.921 &   4144 \\ \hline
      Ask for Confirmation &      0.777 &   0.781 &     0.779 &   2069 \\ \hline
       Does Not Understand &      0.524 &   0.495 &     0.509 &   1417 \\ \hline
       Provides Assistance &      0.884 &   0.902 &     0.893 &   1010 \\ \hline
Ignored Consumer Statement &      0.196 &   0.270 &     0.227 &    226 \\ \hline
         Unable to Resolve &      0.599 &   0.807 &     0.688 &    176 \\ \hline
            Input Rejected &      0.236 &   0.392 &     0.295 &    120 \\ \hline
                   Restart &      0.484 &   0.762 &     0.592 &     80 \\ \hline
              Bad Transfer &      0.250 &   0.286 &     0.267 &     14 \\ \hline \hline
   macro avg &      0.544 &   0.621 &     0.574 & 9256 \\ \hline
weighted avg &      0.799 &   0.784 &     0.790 & 9256 \\ \hline

\end{tabular}
\end{footnotesize}
\end{center}
\end{table*}

\subsection{Using ACQIs to Simplify Bot-Tuning}
\label{sec:systemchanges}

As a final result, we estimate the extent to which predicting the correct ACQI could help 
bot-builders involved in bot-tuning.
Referring back to the ACQI taxonomy in Table \ref{tab:acqi_taxonomy}, without any extra contextual guidance,
bot-builders have 28 possible action strategies with \LP \bots and 31 with \lego \bots.

This was compared with the number of options that would be available using the predict ACQI labeling.
For this simple simulation, we made the following assumptions:

\begin{enumerate}
  \item Each appropriate action is equally likely in the absence of IQ/ACQI. 

  \item If IQ is available, tuning is only required when the score decrements. If IQ is unavailable then all actions related to the system are relevant. 
  
  \item Given the presence of a decremented IQ score, each action is equally likely.
  
  \item If an ACQI is available, all actions that are not assigned to at least one ACQI are included in our list of options that the bot-builder can make. 
  
  \item There is always a special action, \texttt{\small{$\langle$No Action$\rangle$}}, that may applicable for the bot-builder. 
\end{enumerate}

The results of this exercise are presented in Table \ref{tab:perplexity}. 
We found that the largest single simplification comes from the use of IQ --- if IQ can be modelled accurately, then the
average number of recommended options is reduced from 28.6 to 9.73 (about 34\%). Adding ACQI classifications
as well reduces the average down to 5.4 (about 19\% of the original number). This makes a hypothetical but strong 
case: if IQ and ACQI can be accurately predicted on a turn-by-turn level, the amount of effort it takes a bot-builder 
to diagnose problems and suggest possible solutions could be reduced by an estimated 81\%.
While this is an optimistic hypothesis, the potential reward is large enough to encourage more development in this area.

\begin{table*}[t!] 
      \caption{Average number of recommended actions per dialog system when there is no measurement strategy (None), IQ is available (assuming no actions required when score does not decrement), ACQI alone is available, and IQ+ACQI. 95\% confidence intervals were calculated taking 1000 bootstrapped samples (at turn level) per dialog system.}
    \label{tab:perplexity}
    \begin{footnotesize}
      \begin{center}
    \begin{tabular}{|c|c|c|c|c|}
        \hline
       
         \textbf{Dialog System} & \textbf{None} & \textbf{IQ} & \textbf{ACQI} &  \textbf{IQ + ACQI}  \\ 
         \hline\hline
         CMU   & 31  & 7.61 $\pm$ .28     & 7.25 $\pm$ .04   & \textbf{4.2} $\pm$ .14 \\ \hline
         Junior Sales Assistant & 28  &  10.95 $\pm$ .99   & 14.29 $\pm$ .11  & \textbf{5.99} $\pm$ .50    \\ \hline
         Help and Route & 28  & 7.87 $\pm$ .83     & 14.49 $\pm$ .07  & \textbf{4.57} $\pm$ .40  \\ \hline
         Router & 28  &  13.26 $\pm$ 1.43  & 14.77 $\pm$ .16  &  \textbf{7.33} $\pm$ .80  \\ \hline
         Food Expert & 28  & 8.98 $\pm$ .86     & 14.39 $\pm$ .09  & \textbf{4.94} $\pm$ .47 \\ \hline \hline
         All DS & 28.6  & 9.73 $\pm$ .88 & 13.03 $\pm$ .09  & \textbf{5.40} $\pm$ .46 \\ \hline 
    \end{tabular}
          \end{center}
    \end{footnotesize}
\end{table*}

\section{Conclusion}

Actionable Conversation Quality Indicators (ACQIs) are designed to
to provide bot-builders with actionable explanations of why their \bots fail. 
We have explored the key desirable properties for building an ACQI taxonomy, based on recommendations from 
the literature, interviews and collaboration with \bot experts.
Based on an annotated dataset of just over 1000 conversations, we have shown that 
ACQIs are particularly useful when combined with Interaction Quality (IQ), in particular so that
the decision of whether to take a recommended action can be focused on places in the dialog where 
quality decreases. \todo{need to say something about aggregation and how that can mitigate any issues with model performance}

The annotated datasets were used to train predictive models, which achieved a weighted average f1-score of 79\% using
features based just on vectorized embeddings of recent messages in order and logistic regression for classification. 
While these results are preliminary, such a classification model could be used to reduce the number of options for a bot-builder to consider by 
as much as 81\%. Results like this should influence UI design for bot-builders directly: if the ACQI-based suggestions show up 
as a tooltip (similar to refactoring tips in software integrated development environments), they may be useful in the majority 
of cases, while being easy to ignore in the remainder.

The prioritization of the bot-builder as a key user persona is the driving principle for much of this work. 
We hope that more research focused on making bot-builders more effective is encouraged and highlighted in the \bot community,
as a crucial route to optimizing the experience of \bot users overall.

\section*{Competing Interests}

All authors are employed at \LP, Inc. The authors declare no other competing interest.

\bibliographystyle{nlelike}
\bibliography{references}

\begin{thebibliography}{}

\bibitem[Adiwardana et~al., 2020]{adiwardana2020towards}
{\bf Adiwardana, D.}, {\bf Luong, M.-T.}, {\bf So, D.~R.}, {\bf Hall, J.}, {\bf
  Fiedel, N.}, {\bf Thoppilan, R.}, {\bf Yang, Z.}, {\bf Kulshreshtha, A.},
  {\bf Nemade, G.}, {\bf Lu, Y.}, \textbf{and} {\bf others} 2020.
\newblock Towards a human-like open-domain chatbot.
\newblock {\em arXiv preprint arXiv:2001.09977}.

\bibitem[Bodigutla et~al., 2019]{bodigutla2019multi}
{\bf Bodigutla, P.~K.}, {\bf Polymenakos, L.}, \textbf{and} {\bf Matsoukas, S.}
  2019.
\newblock Multi-domain conversation quality evaluation via user satisfaction
  estimation.
\newblock {\em arXiv preprint arXiv:1911.08567}.

\bibitem[Bodigutla et~al., 2020]{bodigutla2020joint}
{\bf Bodigutla, P.~K.}, {\bf Tiwari, A.}, {\bf Vargas, J.~V.}, {\bf
  Polymenakos, L.}, \textbf{and} {\bf Matsoukas, S.} 2020.
\newblock Joint turn and dialogue level user satisfaction estimation on
  multi-domain conversations.
\newblock {\em arXiv preprint arXiv:2010.02495}.

\bibitem[Danieli and Gerbino, 1995]{Danieli_Gerbino1995}
{\bf Danieli, M.} \textbf{and} {\bf Gerbino, E.} 1995.
\newblock Metrics for evaluating dialogue strategies in a spoken language
  system.
\newblock In {\em Proceedings of the 1995 AAAI Spring Symposium on Empirical
  Methods in Discourse Interpretation and Generation}, pp. 34--39.

\bibitem[Deriu et~al., 2020]{deriu2020survey}
{\bf Deriu, J.}, {\bf Rodrigo, A.}, {\bf Otegi, A.}, {\bf Echegoyen, G.}, {\bf
  Rosset, S.}, {\bf Agirre, E.}, \textbf{and} {\bf Cieliebak, M.} 2020.
\newblock Survey on evaluation methods for dialogue systems.
\newblock {\em Artificial Intelligence Review}, pp. 1--56.

\bibitem[Finch and Choi, 2020]{finch2020towards}
{\bf Finch, S.~E.} \textbf{and} {\bf Choi, J.~D.} 2020.
\newblock Towards unified dialogue system evaluation: A comprehensive analysis
  of current evaluation protocols.
\newblock {\em arXiv preprint arXiv:2006.06110}.

\bibitem[{Forrester Research}, 2020]{forrester2020watson}
{\bf {Forrester Research}} 2020.
\newblock The total economic impact of {IBM Watson} assistant.
\newblock Technical report, Forrester, commissioned by IBM.

\bibitem[Han and Anderson, 2020]{han2020customer}
{\bf Han, S.} \textbf{and} {\bf Anderson, C.~K.} 2020.
\newblock Customer motivation and response bias in online reviews.
\newblock {\em Cornell Hospitality Quarterly}, 61(2):142--153.

\bibitem[Hirschman et~al., 1990]{hirschman-etal-1990-beyond}
{\bf Hirschman, L.}, {\bf Dahl, D.~A.}, {\bf McKay, D.~P.}, {\bf Norton,
  L.~M.}, \textbf{and} {\bf Linebarger, M.~C.} 1990.
\newblock Beyond class a: A proposal for automatic evaluation of discourse.
\newblock In {\em Speech and Natural Language: Proceedings of a Workshop Held
  at Hidden Valley, {P}ennsylvania, June 24-27,1990}.

\bibitem[Hirschman and Pao, 1993]{hirshman_pao1993}
{\bf Hirschman, L.} \textbf{and} {\bf Pao, C.} 1993.
\newblock The cost of errors in a spoken language system.
\newblock In {\em Proceedings of the Third European Conference on Speech
  Communication and Technology}, pp. 1419--1422.

\bibitem[Hockey et~al., 2003]{hockey-etal-2003-targeted}
{\bf Hockey, B.~A.}, {\bf Lemon, O.}, {\bf Campana, E.}, {\bf Hiatt, L.}, {\bf
  Aist, G.}, {\bf Hieronymus, J.}, {\bf Gruenstein, A.}, \textbf{and} {\bf
  Dowding, J.} 2003.
\newblock Targeted help for spoken dialogue systems.
\newblock In {\em 10th Conference of the {E}uropean Chapter of the Association
  for Computational Linguistics}, Budapest, Hungary. Association for
  Computational Linguistics.

\bibitem[Jain et~al., 2018]{jain2018evaluating}
{\bf Jain, M.}, {\bf Kumar, P.}, {\bf Kota, R.}, \textbf{and} {\bf Patel,
  S.~N.} 2018.
\newblock Evaluating and informing the design of chatbots.
\newblock In {\em Proceedings of the 2018 Designing Interactive Systems
  Conference}, pp. 895--906.

\bibitem[Krstajic et~al., 2014]{krstajic2014cross}
{\bf Krstajic, D.}, {\bf Buturovic, L.~J.}, {\bf Leahy, D.~E.}, \textbf{and}
  {\bf Thomas, S.} 2014.
\newblock Cross-validation pitfalls when selecting and assessing regression and
  classification models.
\newblock {\em Journal of cheminformatics}, 6(1):1--15.

\bibitem[Landis and Koch, 1977]{landis1977measurement}
{\bf Landis, J.~R.} \textbf{and} {\bf Koch, G.~G.} 1977.
\newblock The measurement of observer agreement for categorical data.
\newblock {\em biometrics}, pp. 159--174.

\bibitem[Li and Sun, 2018]{li2018syntactically}
{\bf Li, J.} \textbf{and} {\bf Sun, X.} 2018.
\newblock A syntactically constrained bidirectional-asynchronous approach for
  emotional conversation generation.
\newblock {\em arXiv preprint arXiv:1806.07000}.

\bibitem[Lin et~al., 2019]{lin2019moel}
{\bf Lin, Z.}, {\bf Madotto, A.}, {\bf Shin, J.}, {\bf Xu, P.}, \textbf{and}
  {\bf Fung, P.} 2019.
\newblock Moel: Mixture of empathetic listeners.
\newblock {\em arXiv preprint arXiv:1908.07687}.

\bibitem[Liu et~al., 2018]{liu2018knowledge}
{\bf Liu, S.}, {\bf Chen, H.}, {\bf Ren, Z.}, {\bf Feng, Y.}, {\bf Liu, Q.},
  \textbf{and} {\bf Yin, D.} 2018.
\newblock Knowledge diffusion for neural dialogue generation.
\newblock In {\em Proceedings of the 56th Annual Meeting of the Association for
  Computational Linguistics (Volume 1: Long Papers)}, pp. 1489--1498.

\bibitem[Luo et~al., 2018]{luo2018auto}
{\bf Luo, L.}, {\bf Xu, J.}, {\bf Lin, J.}, {\bf Zeng, Q.}, \textbf{and} {\bf
  Sun, X.} 2018.
\newblock An auto-encoder matching model for learning utterance-level semantic
  dependency in dialogue generation.
\newblock {\em arXiv preprint arXiv:1808.08795}.

\bibitem[Moghe et~al., 2018]{moghe2018towards}
{\bf Moghe, N.}, {\bf Arora, S.}, {\bf Banerjee, S.}, \textbf{and} {\bf Khapra,
  M.~M.} 2018.
\newblock Towards exploiting background knowledge for building conversation
  systems.
\newblock {\em arXiv preprint arXiv:1809.08205}.

\bibitem[Pieraccini et~al., 2009]{pieraccini2009we}
{\bf Pieraccini, R.}, {\bf Suendermann, D.}, {\bf Dayanidhi, K.}, \textbf{and}
  {\bf Liscombe, J.} 2009.
\newblock Are we there yet? research in commercial spoken dialog systems.
\newblock In {\em International Conference on Text, Speech and Dialogue}, pp.
  3--13. Springer.

\bibitem[Polifroni et~al., 1992]{polifroni1992}
{\bf Polifroni, J.}, {\bf Hirschman, L.}, {\bf Seneff, S.}, \textbf{and} {\bf
  Zue, V.} 1992.
\newblock Experiments in evaluating interactive spoken language systems.
\newblock In {\em Proceedings of the DARPA Speech and NL Workshop}, pp. 28--33.

\bibitem[Qiu et~al., 2019]{qiu2019training}
{\bf Qiu, L.}, {\bf Li, J.}, {\bf Bi, W.}, {\bf Zhao, D.}, \textbf{and} {\bf
  Yan, R.} 2019.
\newblock Are training samples correlated? learning to generate dialogue
  responses with multiple references.
\newblock In {\em Proceedings of the 57th Annual Meeting of the Association for
  Computational Linguistics}, pp. 3826--3835.

\bibitem[Rach et~al., 2017]{rach2017interaction}
{\bf Rach, N.}, {\bf Minker, W.}, \textbf{and} {\bf Ultes, S.} 2017.
\newblock Interaction quality estimation using long short-term memories.
\newblock In {\em Proceedings of the 18th Annual SIGdial Meeting on Discourse
  and Dialogue}, pp. 164--169.

\bibitem[Raux et~al., 2006]{Raux2006DoingRO}
{\bf Raux, A.}, {\bf Bohus, D.}, {\bf Langner, B.}, {\bf Black, A.},
  \textbf{and} {\bf Esk{\'e}nazi, M.} 2006.
\newblock Doing research on a deployed spoken dialogue system: one year of
  let's go! experience.
\newblock In {\em INTERSPEECH}.

\bibitem[Raux et~al., 2005]{Raux2005LetsGP}
{\bf Raux, A.}, {\bf Langner, B.}, {\bf Bohus, D.}, {\bf Black, A.},
  \textbf{and} {\bf Esk{\'e}nazi, M.} 2005.
\newblock Let's go public! taking a spoken dialog system to the real world.
\newblock In {\em INTERSPEECH}.

\bibitem[Reimers and Gurevych, 2019]{reimers2019sentence}
{\bf Reimers, N.} \textbf{and} {\bf Gurevych, I.} 2019.
\newblock Sentence-bert: Sentence embeddings using siamese bert-networks.
\newblock {\em arXiv preprint arXiv:1908.10084}.

\bibitem[Rozin and Royzman, 2001]{negativity_bias}
{\bf Rozin, P.} \textbf{and} {\bf Royzman, E.~B.} 2001.
\newblock Negativity bias, negativity dominance, and contagion.
\newblock {\em Personality and social psychology review}, 5(4):296--320.

\bibitem[Schmitt et~al., 2011]{schmitt2011modeling}
{\bf Schmitt, A.}, {\bf Schatz, B.}, \textbf{and} {\bf Minker, W.} 2011.
\newblock Modeling and predicting quality in spoken human-computer interaction.
\newblock In {\em Proceedings of the SIGDIAL 2011 Conference}, pp. 173--184.

\bibitem[Schmitt and Ultes, 2015]{schmitt2015interaction}
{\bf Schmitt, A.} \textbf{and} {\bf Ultes, S.} 2015.
\newblock Interaction quality: assessing the quality of ongoing spoken dialog
  interaction by experts—and how it relates to user satisfaction.
\newblock {\em Speech Communication}, 74:12--36.

\bibitem[Schmitt et~al., 2012]{schmitt2012parameterized}
{\bf Schmitt, A.}, {\bf Ultes, S.}, \textbf{and} {\bf Minker, W.} 2012.
\newblock A parameterized and annotated spoken dialog corpus of the {CMU} let's
  go bus information system.
\newblock In {\em LREC}, pp. 3369--3373.

\bibitem[Shriberg et~al., 1992]{shriberg1992}
{\bf Shriberg, E.}, {\bf Wade, E.}, \textbf{and} {\bf Price, P.} 1992.
\newblock Human-machine problem solving using spoken language systems (sls):
  Factors affecting performance and user satisfaction.
\newblock In {\em Proceedings of the DARPA Speech and NL Workshop}, pp. 49--54.

\bibitem[Stoyanchev et~al., 2017]{Stoyanchev2017PredictingIQ}
{\bf Stoyanchev, S.}, {\bf Maiti, S.}, \textbf{and} {\bf Bangalore, S.} 2017.
\newblock Predicting interaction quality in customer service dialogs.
\newblock In {\em IWSDS}.

\bibitem[Stoyanchev et~al., 2019]{stoyanchev2019predicting}
{\bf Stoyanchev, S.}, {\bf Maiti, S.}, \textbf{and} {\bf Bangalore, S.} 2019.
\newblock Predicting interaction quality in customer service dialogs.
\newblock In {\em Advanced Social Interaction with Agents}, pp. 149--159.
  Springer.

\bibitem[Szyma{\'n}ski and Kajdanowicz, 2017]{szymanski2017network}
{\bf Szyma{\'n}ski, P.} \textbf{and} {\bf Kajdanowicz, T.} 2017.
\newblock A network perspective on stratification of multi-label data.
\newblock In {\em First International Workshop on Learning with Imbalanced
  Domains: Theory and Applications}, pp. 22--35. PMLR.

\bibitem[Ultes et~al., 2015]{LEGOv2}
{\bf Ultes, S.}, {\bf S{\'a}nchez, M. J.~P.}, {\bf Schmitt, A.}, \textbf{and}
  {\bf Minker, W.} 2015.
\newblock Analysis of an extended interaction quality corpus.
\newblock In {\em Natural Language Dialog Systems and Intelligent Assistants},
  pp. 41--52. Springer.

\bibitem[Walker et~al., 1997]{paradise}
{\bf Walker, M.~A.}, {\bf Litman, D.~J.}, {\bf Kamm, C.~A.}, \textbf{and} {\bf
  Abella, A.} 1997.
\newblock {PARADISE}: A framework for evaluating spoken dialogue agents.
\newblock In {\em 35th Annual Meeting of the Association for Computational
  Linguistics and 8th Conference of the {E}uropean Chapter of the Association
  for Computational Linguistics}, pp. 271--280, Madrid, Spain. Association for
  Computational Linguistics.

\bibitem[Wang et~al., 2020]{wang2020improving}
{\bf Wang, J.}, {\bf Liu, J.}, {\bf Bi, W.}, {\bf Liu, X.}, {\bf He, K.}, {\bf
  Xu, R.}, \textbf{and} {\bf Yang, M.} 2020.
\newblock Improving knowledge-aware dialogue generation via knowledge base
  question answering.
\newblock In {\em Proceedings of the AAAI Conference on Artificial
  Intelligence}, pp. 9169--9176.

\bibitem[Witt, 2011]{witt2011global}
{\bf Witt, S.} 2011.
\newblock A global experience metric for dialog management in spoken dialog
  systems.
\newblock In {\em Proceedings of SemDial}, pp. 158--166.

\bibitem[Wu et~al., 2019]{wu2019proactive}
{\bf Wu, W.}, {\bf Guo, Z.}, {\bf Zhou, X.}, {\bf Wu, H.}, {\bf Zhang, X.},
  {\bf Lian, R.}, \textbf{and} {\bf Wang, H.} 2019.
\newblock Proactive human-machine conversation with explicit conversation
  goals.
\newblock {\em arXiv preprint arXiv:1906.05572}.

\end{thebibliography}

\end{document}